\documentclass[runningheads]{llncs}
\usepackage[T1]{fontenc}
\usepackage{graphicx}
\usepackage{amssymb}
\usepackage{multirow}
\usepackage{booktabs}
\usepackage{hyperref}
\usepackage{color}

\begin{document}
\title{Robust and Efficient Writer-Independent IMU-Based Handwriting Recognition}
\titlerunning{Robust and Efficient WI IMU-Based HWR}
%
\author{Jindong Li\inst{1}
    \orcidID{0000-0002-3550-1660} \and
    Tim Hamann\inst{2}\orcidID{0000-0003-3562-6882} \and
    Jens Barth\inst{2}\orcidID{0000-0003-3967-9578} \and
    Peter Kämpf\inst{2} \and
    Dario Zanca\inst{1}\orcidID{0000-0001-5886-0597} \and
    Björn Eskofier\inst{1, 3}\orcidID{0000-0002-0417-0336}}
\authorrunning{J. Li et al.}
%
\institute{Machine Learning and Data Analytics Lab, Friedrich-Alexander-Universität Erlangen-Nürnberg, Erlangen, Germany \\
    \and STABILO International GmbH, Heroldsberg, Germany \and Translational Digital Health Group, Institute of AI for Health, Helmholtz Zentrum München - German Research Center for Environmental Health, Neuherberg, Germany}
\maketitle              
\begin{abstract}

    Handwriting recognition (HWR) using inertial measurement unit (IMU) data remains challenging due to variations in writing styles and the limited availability of datasets. Previous approaches often struggle with handwriting from unseen writers, making writer-independent (WI) recognition a crucial yet difficult problem. This paper presents a model designed to improve WI HWR on IMU data, using a CNN encoder and BiLSTM-based decoder. Our approach demonstrates strong robustness to unseen handwriting styles, outperforming existing methods on the WI splits of both the public OnHW dataset and our word-based dataset, achieving character error rates (CERs) of 7.37\% and 9.44\%, and word error rates (WERs) of 15.12\% and 32.17\%, respectively. Robustness evaluation shows that our model maintains superior performance across different age groups, with knowledge learned from one group generalizing better to another compared to other approaches. Evaluation on our sentence-based dataset further demonstrates the potential for recognizing full sentences. Through comprehensive ablation studies, we show that our design choices achieve a strong balance between performance and efficiency. These findings support the development of more adaptable and scalable HWR systems for real-world applications. The code is available at: \url{https://github.com/jindongli24/REWI}.

\end{abstract}
\section{Introduction}

Handwriting has served as a primary means of recording and sharing information throughout human history. As technology has advanced, the demand for digitizing handwritten content has grown significantly. HWR, which converts handwritten symbols into computer-readable text, has consequently emerged as an important area of research.

Offline HWR, as one of the main types of HWR, has been widely used in various fields\cite{ocr_historical,ocr_medical}. However, it is limited to recognizing static images of handwritten text. In contrast, online HWR processes time-series data in real-time, capturing dynamic handwriting features such as stroke trajectories, tip positions, writing directions, and velocities as they occur. This data is typically collected using touchscreens and styluses on mobile devices \cite{ohwr_stylus}, which limits the available writing surface for users. Another approach to online HWR uses pens equipped with IMUs \cite{digipen_stabilo,digipen_alemayoh,digipen_meissl}. These sensors capture pen movements without requiring precise tip positioning, enabling the pen to operate independently of external devices and demonstrating strong potential for broader online HWR applications.

However, IMU-based HWR faces several challenges that affect recognition performance. First, similar to other HWR approaches, handwritten text varies vastly across different writers in terms of style, including slant angle, letter connections, and spacing between letters and words, making WI recognition particularly challenging \cite{wsa}. Additionally, noise from rough surfaces, signal drift caused by heat accumulation inside the device, and artifacts during signal communication or transmission present additional technical difficulties \cite{ohwr_stabilo_2}. Moreover, as an online task, latency must also be considered, which poses another challenge for efficiency.

This paper presents a sequence-to-sequence model for IMU-based online HWR, designed to address efficiency challenges and those arising from handwriting style variations and noise. The model adopts an encoder-decoder architecture that combines a convolutional neural network (CNN) with a bidirectional long short-term memory (BiLSTM) network. We evaluate the proposed model by benchmarking it against several existing IMU-based HWR methods, as well as adapted mainstream models that have achieved strong performance in various deep learning tasks. Evaluation is performed on the public OnHW dataset and our custom IMU-based handwriting datasets, where our datasets each address different aspects of the problem space. Experimental results demonstrate that our model outperforms others in terms of efficiency, robustness, and flexibility for WI HWR.

\section{Related Works}
\subsection{IMU-based HWR}

IMU-based HWR has been an active area of research for decades. Early studies, such as \cite{imu_digit_dtw_1,imu_digit_dtw_2}, employed dynamic time warping algorithms to recognize digit data collected with IMU-based pens, achieving recognition rates above 90\%. Later works, including \cite{imu_char,digipen_meissl}, explored LSTM-based models for recognizing individual English characters, reporting recognition accuracies of up to 99.68\% and 79.01\% on their respective datasets. Although these methods achieved excellent performance, they rely on isolated character recognition, processing entire input sequences to classify single characters. This character-by-character approach disrupts the natural writing flow, making it impractical for real-world applications where writing typically occurs word by word.

To address more complex tasks, \cite{ohwr_stabilo_1,ohwr_stabilo_2} employed CNN-BiLSTM models with connectionist temporal classification (CTC) \cite{ctc} to recognize sequences of English and German characters. Additionally, \cite{ohwr_stabilo_2} explored the potential of various architectures for IMU-based HWR. These models were evaluated on datasets collected using the IMU-based pen developed by STABILO, achieving CERs of 27.8\% and 17.97\%, respectively. These results marked an important step toward recognizing full words and sentences, bringing the field closer to practical applications.

\subsection{Advancements in Deep Learning Architectures}

In recent years, the introduction of ResNet \cite{resnet} and Transformer \cite{transformer} has significantly advanced deep learning model development. ResNet addressed the vanishing gradient problem through skip connections, enabling effective training of very deep neural networks. Transformers introduced self-attention mechanisms, which improved parallelization and enhanced the modeling of long-range dependencies compared to CNNs or recurrent neural networks.

Building on these innovations, several new approaches have emerged. MLP-Mixer \cite{mlp_mixer} demonstrated that strong performance in vision tasks can be achieved without convolutions or self-attention, relying solely on multi-layer perceptrons to mix spatial and channel information. Swin Transformers \cite{swinv1,swinv2} introduced a hierarchical architecture with shifted window-based self-attention, enabling efficient multi-scale modeling for various vision tasks. Vision Transformer (ViT) \cite{vit} treats images as sequences of patch embeddings fed into a Transformer encoder, showing that pure self-attention can rival CNNs on large-scale vision tasks. ConvNeXt \cite{convnext} integrated concepts from both CNNs and Transformers, achieving state-of-the-art performance while preserving the simplicity and efficiency of traditional CNNs.

Although these models were originally designed for different tasks, their core principles, such as self-attention and hierarchical structures, are potentially well-suited for time-series data. Applying these techniques to IMU-based HWR could enhance performance and open new research opportunities in the field.

\section{Methods}
\subsection{Datasets}
\subsubsection{Private Dataset}

Our datasets were collected using the IMU-based pen developed by STABILO. Data collection followed pseudo-anonymous procedures where voluntary participants' names appeared only on consent agreements, while all data were stored anonymously on the server, ensuring that only minimal and necessary personal information was collected. For commercial reasons, our private dataset will not be made publicly available.

\begin{figure}[htbp]
    \centering
    \includegraphics[width=\linewidth]{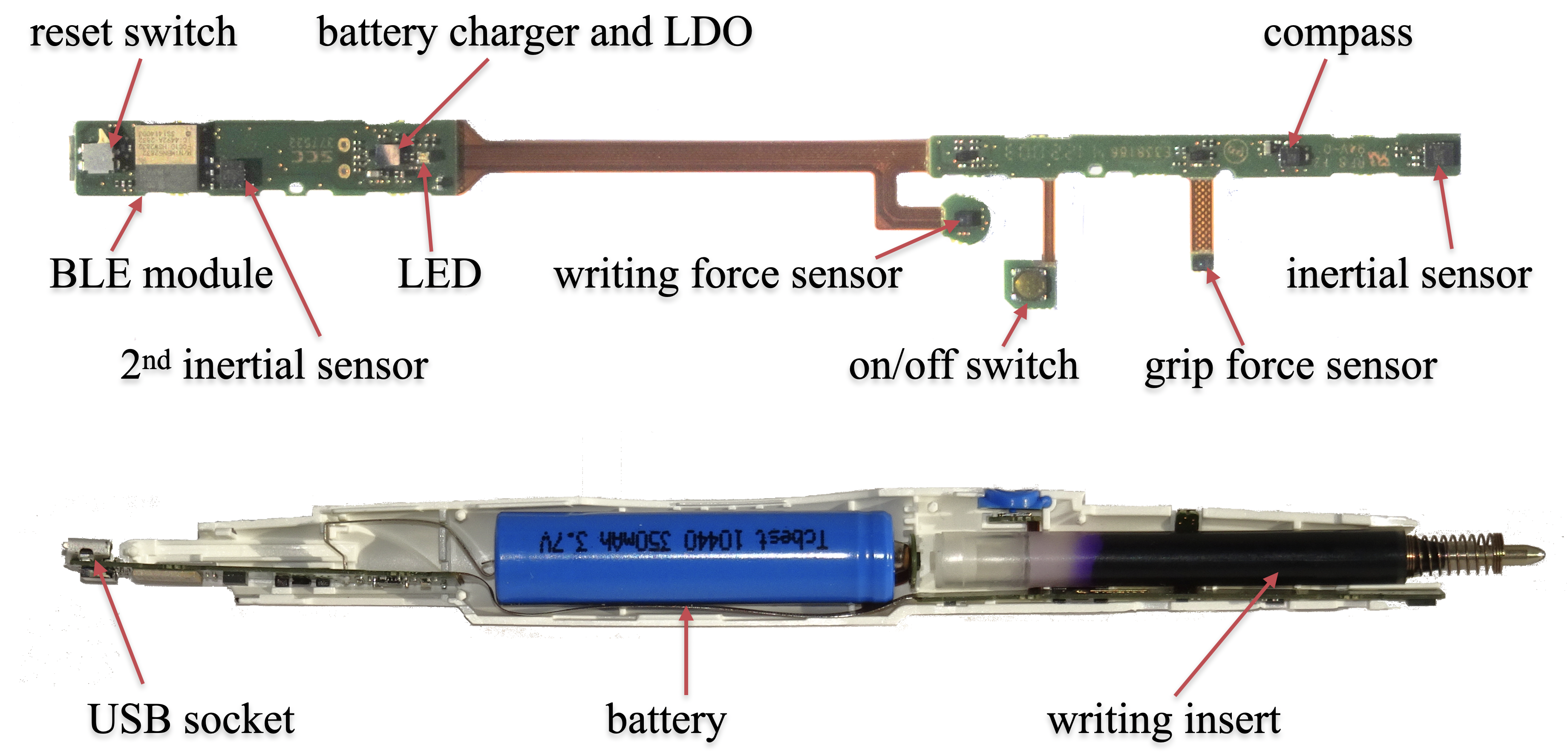}
    \caption{\textbf{Digipen Architecture.} The pen includes two IMUs (one with an accelerometer and gyroscope, the other with only an accelerometer) located at both ends, a magnetometer (compass), a writing force sensor, and a grip force sensor (not used in the experiments).}
    \label{fig:digipen}
\end{figure}

The pen generates 13 output channels at a selectable sampling rate of 100 Hz, 200 Hz, or 400 Hz, using two accelerometers (one at each end), a gyroscope, a magnetometer, and a force sensor, as shown in Figure~\ref{fig:digipen}. Each sensor provides three channels of data, except the force sensor, which provides a single channel.

We conducted 1,547 recording sessions with 520 participants of various ages and handedness, collecting both English and German words and sentences. All samples were converted to a 100 Hz sampling rate and divided into two subsets: a word-based dataset and a sentence-based dataset.

The word-based dataset contains 24,163 samples of English and German words, covering 59 character categories, including both uppercase and lowercase letters from both languages. The sentence-based dataset includes 20,639 samples of English and German sentences, spanning 84 character categories, including letters, numbers, symbols, and spaces. Given the distribution of sample lengths, only samples shorter than 1,024 time steps for word signals and 4,096 time steps for sentence signals are retained. Visualizations of character category distributions and sample lengths are provided in Appendix~\ref{app:distribution_character_length}.

After preprocessing, the data samples are split into training (80\%) and validation (20\%) sets using 5-fold cross-validation. The data are divided at the writer level to ensure that no writer appears in both training and validation sets. To maintain balanced representation between the sets, stratification based on participants' handedness and age is applied.

\begin{figure}[htbp]
    \centering
    \includegraphics[width=0.9\linewidth]{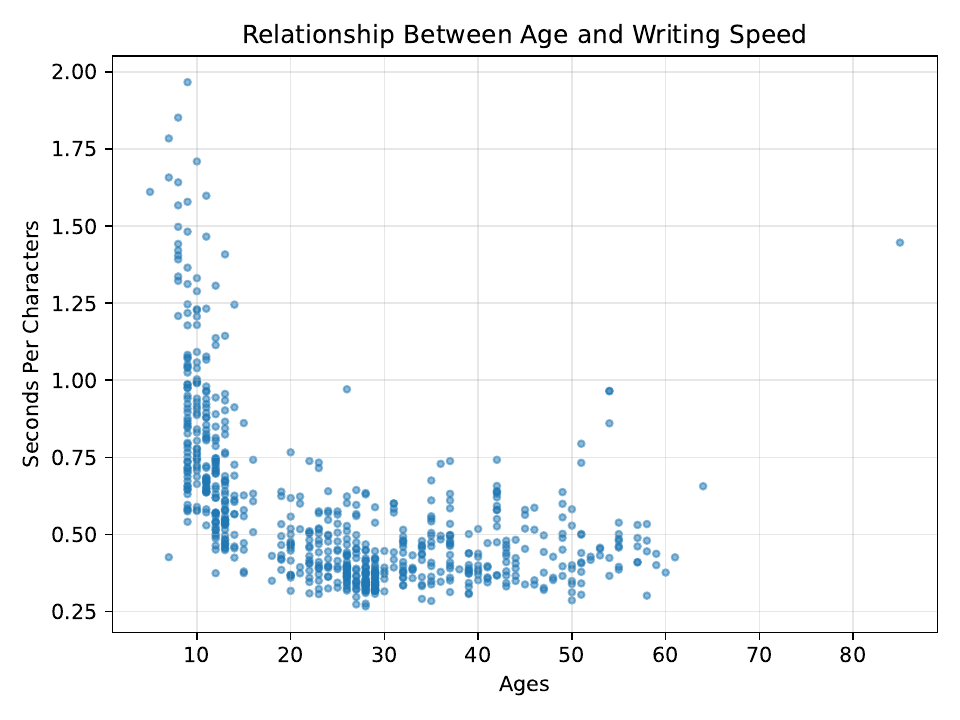}
    \caption{\textbf{Relationship Between Age and Writing Speed.} This visualization shows the average writing time per character for writers in our word-based dataset, illustrating how writing speed varies across different ages.}
    \label{fig:writing_speed}
\end{figure}

To evaluate the robustness of the system, we also analyze subsets of our dataset based on participants' ages. Since children are still developing their handwriting skills, their handwriting patterns differ significantly from those of adults, particularly in writing speed and discontinuity due to high cognitive load, with these patterns typically stabilizing around 14 years old \cite{hw_adult_children_1,hw_adult_children_2}. This finding is further supported by Figure~\ref{fig:writing_speed} and Appendix~\ref{app:distribution_character_length}, which demonstrate that writers under 14 years old write significantly more slowly than their older counterparts.

Based on the distinct writing characteristics of writers at different ages, we classify participants aged 12 and under as children to ensure that most writers represent typical children handwriting patterns, and those aged 18 and older as adults. Data from participants aged 13 to 17 are excluded from these subsets because some have already developed mature and fluent handwriting, making them less representative of either the children or adult groups. Notably, the adult subset (13,899 samples) is approximately twice the size of the children subset (7,762 samples).

\subsubsection{OnHW Dataset}

We also use the WI split from the right-handed subset of the public OnHW-words500 dataset \cite{ohwr_stabilo_2}, which contains 13-channel handwriting data for 500 words written by 53 writers. The left-handed subset, which includes only 1,000 samples and offers only a single train-validation split, is excluded from our experiments due to its limited size and configuration.

\subsection{Model Architecture}

\begin{figure}[htbp]
    \centering
    \includegraphics[width=\linewidth]{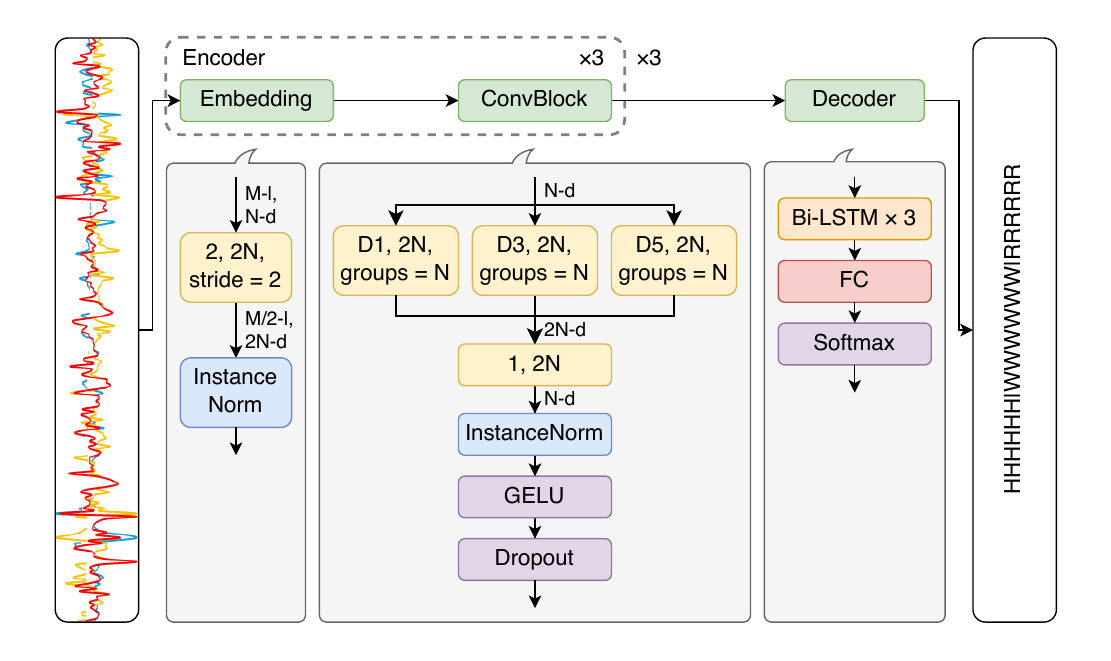}
    \caption{\textbf{Model architecture.} The model consists of a three-stage CNN encoder and a BiLSTM-based decoder. The convolutional blocks use separable depthwise convolutional layers with inverted bottleneck and multi-scale designs to improve efficiency.}
    \label{fig:arch_blcnn}
\end{figure}

Although \cite{ohwr_stabilo_1,ohwr_stabilo_2} have presented promising CNN-BiLSTM-based approaches to HWR of words and delivered strong performance, both works still leave considerable room for improvement. In terms of robustness, neither approach targets unseen handwriting styles, and both struggle with WI HWR. Regarding efficiency, both employ simple and regular CNN designs that cannot effectively utilize the limited number of parameters. To design a robust and efficient IMU-based HWR baseline, building upon CLDNN \cite{ohwr_stabilo_1}, we designed our model by incorporating various delicated modifications in the CNN architecture to improve both efficiency and robustness, as illustrated in Figure~\ref{fig:arch_blcnn}.

\subsubsection{Encoder}

The encoder consists of three stages, each containing an embedding layer followed by three convolutional blocks. Each embedding layer uses a convolutional layer with kernel size 2 and stride 2 to downsample the input sequences, followed by an instance normalization layer \cite{inst_norm} to reduce representation variance between samples with different handwriting styles. Except for the first stage, which processes inputs with 13 channels, each convolutional layer doubles the number of output channels relative to the input channels.

To reduce computational cost, we used an inverted bottleneck design. This starts with a grouped convolutional layer (kernel size 5) that doubles the channel count using one group per input channel. A pointwise convolutional layer then restores the original channel dimensions. This approach creates a larger hidden space for better feature representation while using fewer computational resources than conventional convolutional layers. It also offers advantages over standard depthwise separable convolutions.

Additionally, we introduce a multi-scale design that adds two more grouped convolutional layers with kernel sizes of 1 and 3 during training, in addition to the existing layer with kernel size 5. The input sequence is processed in parallel by all three convolutional layers, and their outputs are summed before passing to the pointwise convolutional layer. After training, we merge the weights from the smaller-kernel layers into the kernel size 5 layer, resulting in the same number of parameters as the original variant of the depthwise separable convolution. This approach enables multi-scale feature learning during training without adding extra parameters or computational cost during inference.

\subsubsection{Decoder}

The decoder consists of three BiLSTM layers, each with a hidden size of 128 and a dropout rate of 0.2. These are followed by a fully connected layer and a softmax layer for pointwise classification, omitting the intermediate hidden layers used in CLDNN to improve efficiency. The softmax outputs are then processed by a greedy CTC decoder to generate the final text predictions.

\subsubsection{Edge Optimization}

To better support edge computing, we also introduce a small version of the model that uses only one convolutional block per stage and reduces the embedded channel sizes from 128, 256, and 512 to 64, 128, and 256. The BiLSTM decoder is also reduced in size, with the hidden size decreased to 64 and the number of layers reduced to 2.

\subsection{Data Augmentation}

To enhance robustness against noise and diverse handwriting styles, we applied four data augmentation schemes: AddNoise, Drift, Dropout, and TimeWarp. Each scheme has a 25\% probability of being applied to a given signal. TimeWarp affects only the time dimension, while the other methods are applied multiplicatively to ensure their contributions are not overshadowed by the magnitude of the original signals. Visualizations of all data augmentation methods are provided in Appendix~\ref{app:data_augmentation}.

\textbf{AddNoise} adds Gaussian noise to the original signals, simulating noise from rough writing surfaces and other sources. \textbf{Drift} divides the signals into segments and applies random drift within each segment, simulating drift caused by heat accumulation inside the pen. \textbf{Dropout} randomly replaces small segments of the signal with the last value preceding each segment, simulating communication imperfections. \textbf{TimeWarp} randomly adjusts the speed of segments within the original signal, simulating variations in writing speed.

After augmentation, each signal sample is individually normalized at the channel level, rather than globally, to minimize the influence of writer-specific characteristics such as pen pressure and pen angle. This approach encourages the model to focus on the actual writing content rather than individual writing habits.

\section{Experiments}

The experiments consist of four parts: evaluation on word-based datasets, robustness evaluation, evaluation on our sentence-based dataset, and an ablation study. In each part, models are trained using 5-fold cross-validation, and the best result from validation set of each fold is used to calculate the average across all five folds, which is reported as the final result.

We first include two previous works: CNN+BiLSTM \cite{ohwr_stabilo_2} and CLDNN \cite{ohwr_stabilo_1}. Since their code is not publicly available, we re-implemented both models in PyTorch based on the original papers. For CLDNN, we followed the described training and preprocessing pipeline. However, training details for CNN+BiLSTM were not clearly specified, so we trained it using our own pipeline to ensure fair comparison. These two models are compared with the small version of our model, given their similar sizes.

For broader comparison, we implemented ResNet, MLP-Mixer, ViT, ConvNeXt, and Swin Transformer V2 (SwinV2) backbones as encoders, replacing the CNN encoder in the base version of our model. Since these backbones were not originally designed for time-series inputs, we made minimal but necessary modifications to them. Due to architectural limitations, MLP-Mixer, ViT, and SwinV2 require fixed-size inputs. Therefore, input sequences are zero-padded to a length of 1,024 for word samples and 4,096 for sentence samples when using these models. Additionally, encoders and decoders are configured with hyperparameters to ensure comparable parameter counts when input sequences are of the same length. Given their sizes, these approaches are compared against the base version of our model.

Except for CLDNN \cite{ohwr_stabilo_1}, which is trained as described in its original paper, all other models are trained using the same procedure. We use the AdamW optimizer with a learning rate of 0.001 for 300 epochs. To improve training stability and convergence, linear learning rate warm-up is applied, starting at 0.0001 for the first 30 epochs, followed by a cosine annealing schedule for the remaining epochs. The training process uses the CTC loss function and a batch size of 64. Models are evaluated every 5 epochs during training.

To measure model performance, we use two metrics: CER and WER. CER reflects errors at the character level, while WER reflects errors at the word level. These metrics measure how many errors appear in the predictions, including substitutions, deletions, and insertions, compared to the total number of characters or words in the ground truth text. We also evaluate memory usage and computational cost by reporting the number of parameters (\#Params) and the number of multiply-accumulate operations (MACs). These are calculated using a random input sequence with 13 channels and 1,024 timesteps for word-based datasets, and 4,096 timesteps for the sentence-based dataset.

\subsection{Evaluation on Word-Based Datasets}

In this section, we compare our models with other approaches using the WI splits of our word-based dataset and the right-handed subset of the OnHW dataset. For reference, we also include the published results for CNN+BiLSTM on the OnHW-words500 dataset \cite{ohwr_stabilo_2}.

\begin{table}[ht]
    \centering
    \addtolength{\tabcolsep}{3pt}
    \caption{\textbf{Results on Our Word-Based Dataset and the Right-Handed Subset of the OnHW Dataset.} We compare our models with other approaches on the WI splits of our dataset and the right-handed subset of the OnHW-words500 dataset.  Metrics include CER, WER, \#Params, and MACs. Error rates are reported as percentages. The best results are shown in \textbf{bold}.}
    \begin{tabular}{lcccccc}
        \toprule
        \multirow{2}{*}{Models} & \multicolumn{2}{c}{Ours} & \multicolumn{2}{c}{OnHW} & \multirow{2}{*}{\#Params} & \multirow{2}{*}{MACs}                                  \\
                                & CER                      & WER                      & CER                       & WER                   &                &               \\
        \midrule
        CLDNN                   & 19.05                    & 60.20                    & 15.62                     & 36.71                 & 0.75M          & 291M          \\
        CNN+BiLSTM (orig.)      & \textemdash              & \textemdash              & 27.80                     & 60.91                 & \textemdash    & \textemdash   \\
        CNN+BiLSTM              & 21.32                    & 64.83                    & 17.66                     & 43.45                 & \textbf{0.40M} & 152M          \\
        Ours-Small              & \textbf{15.29}           & \textbf{50.70}           & \textbf{10.55}            & \textbf{24.94}        & 0.53M          & \textbf{79M}  \\
        \midrule
        ResNet (enc.)           & 10.89                    & 36.71                    & 8.22                      & 18.35                 & 3.97M          & 591M          \\
        MLP-Mixer (enc.)        & 13.37                    & 43.25                    & 9.74                      & 21.87                 & 3.90M          & 802M          \\
        ViT (enc.)              & 13.89                    & 45.24                    & 10.49                     & 22.71                 & \textbf{3.71M} & \textbf{477M} \\
        ConvNeXt (enc.)         & 12.25                    & 42.16                    & 8.23                      & 18.46                 & 3.86M          & 600M          \\
        SwinV2 (enc.)           & 11.70                    & 40.69                    & 8.62                      & 19.60                 & 3.88M          & 601M          \\
        Ours-Base               & \textbf{9.44}            & \textbf{32.17}           & \textbf{7.37}             & \textbf{15.12}        & 3.89M          & 600M          \\
        \bottomrule
    \end{tabular}
    \label{tab:evaluation_word}
\end{table}

As shown in Table 1, the small version of our model outperforms other IMU-based HWR approaches in both CER and WER on our dataset (15.29\% and 50.70\%) and the OnHW dataset (10.55\% and 24.94\%), while maintaining the lowest computational cost. The second-best approach, CLDNN, performs 24.6\% and 18.7\% worse on our dataset, and 48.1\% and 47.2\% worse on the OnHW dataset, respectively. Furthermore, CLDNN has 43.4\% more parameters and 268\% higher computational cost. When comparing the CER and WER of CNN+BiLSTM from the original paper with those from our re-implementation, our training pipeline and data augmentation strategies clearly lead to significant performance improvements, reducing CER and WER by 36.5\% and 28.7\%, respectively.

Compared to other competitors, the base version of our model achieves the lowest CER and WER on our dataset (9.44\% and 32.17\%) and the OnHW dataset (7.37\% and 15.12\%). Notably, unlike backbones designed for local attention, such as CNNs and SwinV2, backbones that focus on global information, like ViT and MLP-Mixer, perform poorly across all metrics. This reveals that IMU-based HWR prioritizes local information over global information.

Additionally, we investigate the errors made by our base-version model on our word-based dataset. As shown in Appendix~\ref{app:substitution_error_matrix}, substitution is the primary error type. Most substitution errors result from confusion between uppercase and lowercase letters with similar patterns, such as "c"/"C" and "w"/"W". Similarly, there are also instances of confusion between different letters with similar shapes, such as "q" and "g".

\subsection{Robustness Evaluation}

In this section, we assess the robustness of the models by evaluating them on the WI splits of the adult and children subsets of our word-based dataset. Additionally, we test the performance of models trained on the adult subset against the validation sets of the children subset (Adult2Child), and vice versa (Child2Adult). Trained models from each fold are tested on all validation sets of the other subset, and the mean of all 25 evaluation results is reported as the final result.

\begin{table}[ht]
    \centering
    \addtolength{\tabcolsep}{3pt}
    \caption{\textbf{Results on the Adult and Children Subsets of Our Word-Based Dataset.} We compare our models with other approaches on the WI splits of the adult and children subsets of our word-based dataset, categorized by age group (Adults: $\geqslant$ 18 years old; Children: $\leqslant$ 12 years old). Models trained on one subset are also evaluated on the other (Adult2Child / Child2Adult).}
    \begin{tabular}{lcccccccc}
        \toprule
        \multirow{2}{*}{Models} & \multicolumn{2}{c}{Adults} & \multicolumn{2}{c}{Children} & \multicolumn{2}{c}{Adult2Child} & \multicolumn{2}{c}{Child2Adult}                                                                     \\
                                & CER                        & WER                          & CER                             & WER                             & CER            & WER            & CER            & WER            \\
        \midrule
        CLDNN                   & 21.62                      & 63.79                        & 28.78                           & 74.06                           & 44.67          & 86.74          & 52.32          & 94.62          \\
        CNN+BiLSTM              & 22.99                      & 67.31                        & 32.11                           & 78.48                           & 45.97          & 87.25          & 47.08          & 92.61          \\
        Ours-Small              & \textbf{16.22}             & \textbf{52.40}               & \textbf{22.97}                  & \textbf{65.47}                  & \textbf{41.28} & \textbf{81.48} & \textbf{40.69} & \textbf{87.96} \\
        \midrule
        ResNet (enc.)           & 13.74                      & 42.96                        & 18.07                           & 53.64                           & 37.94          & 78.59          & 37.89          & 85.43          \\
        MLP-Mixer (enc.)        & 16.30                      & 50.89                        & 75.29                           & 98.35                           & 41.33          & 81.94          & 79.98          & 99.74          \\
        ViT (enc.)              & 16.42                      & 51.32                        & 32.71                           & 74.88                           & 39.73          & 81.02          & 49.74          & 93.35          \\
        ConvNeXt (enc.)         & 14.63                      & 47.07                        & 24.85                           & 63.85                           & 37.11          & 79.17          & 44.48          & 90.23          \\
        SwinV2 (enc.)           & 14.49                      & 47.48                        & 24.93                           & 66.13                           & 35.97          & 78.63          & 42.43          & 89.53          \\
        Ours-Base               & \textbf{10.98}             & \textbf{36.16}               & \textbf{15.00}                  & \textbf{46.11}                  & \textbf{35.86} & \textbf{74.86} & \textbf{36.22} & \textbf{82.88} \\
        \bottomrule
    \end{tabular}
    \label{tab:robustness_evaluation}
\end{table}

As shown in Table~\ref{tab:robustness_evaluation}, on the adult subset, our small-version model achieves CER and WER that are 25.0\% and 17.9\% better than the second-best model, CLDNN. Our base-version model outperforms its closest competitor, ResNet, by 20.1\% and 15.8\% in CER and WER, respectively. On the children subset, CLDNN and ResNet remain the strongest among the competing models. However, CLDNN is still 25.3\% and 13.1\% worse than our small-version model in CER and WER, while ResNet is 20.5\% and 16.3\% worse than our base-version model.

For cross-subset evaluation, our models consistently achieve the best performance across all evaluations. When trained on the larger adult subset and evaluated on the children subset, our small-version model performs 7.6\% and 6.1\% better than its best-performing competitor, CLDNN, in CER and WER, while our base-version model performs 17.0\% and 14.0\% better than ResNet. Conversely, when trained on the children subset and evaluated on the adult subset, our small-version model performs 13.6\% and 5.0\% better than CNN+BiLSTM, while our base-version model performs 4.4\% and 3.0\% better than ResNet.

Notably, the MLP-Mixer failed to converge when trained on the children subset, resulting in poor performance across both evaluation subsets.

We also compare the error patterns made by our base-version model on the adult and children subsets when trained on their respective datasets. As shown in Appendix~\ref{app:substitution_error_matrix}, the children subset presents significantly more challenging conditions. While the adult subset exhibits sparse, predictable errors primarily between visually similar characters or rare characters, the children subset shows dense, widespread confusion patterns across multiple character classes that extend beyond shape similarity. This degraded performance may be attributed partly to the children subset being smaller than the adult subset, but also to the additional recognition complexities inherent in children handwriting.

\subsection{Evaluation on Our Sentence-Based Dataset}

In this section, we compare our models with other approaches using the WI split of our sentence-based dataset.

\begin{table}[ht]
    \centering
    \addtolength{\tabcolsep}{3pt}
    \caption{\textbf{Results on Our Sentence-Based Dataset.} We compare our models with other approaches on the WI split of our sentence-based dataset.}
    \begin{tabular}{lcccc}
        \toprule
        Models           & CER            & WER            & \#Params       & MACs           \\
        \midrule
        CLDNN            & 17.47          & 59.17          & 0.75M          & 1.17B          \\
        CNN+BiLSTM       & 18.94          & 66.25          & \textbf{0.41M} & 0.61B          \\
        Ours-Small       & \textbf{13.22} & \textbf{48.75} & 0.54M          & \textbf{0.32B} \\
        \midrule
        ResNet (enc.)    & 8.20           & 28.74          & 3.97M          & 2.37B          \\
        MLP-Mixer (enc.) & 75.14          & 100.00         & 15.71M         & 8.04B          \\
        ViT (enc.)       & 10.57          & 37.35          & \textbf{3.72M} & \textbf{1.93B} \\
        ConvNeXt (enc.)  & 9.48           & 34.72          & 3.87M          & 2.40B          \\
        SwinV2 (enc.)    & 9.06           & 33.44          & 3.89M          & 2.41B          \\
        Ours-Base        & \textbf{6.78}  & \textbf{24.63} & 3.90M          & 2.40B          \\
        \bottomrule
    \end{tabular}
    \label{tab:evaluation_ours_sentence}
\end{table}

As shown in Table~\ref{tab:evaluation_ours_sentence}, our models achieve the lowest error rates. The small version reaches a CER of 13.22\% and a WER of 48.75\%, which are 24.3\% and 17.6\% better than the second-best model, CLDNN, while also consuming the least computational resources. The base version achieves a CER of 6.78\% and a WER of 24.63\%, outperforming the second-best model, ResNet, by 20.9\% and 16.7\%, respectively.

Notably, the MLP-Mixer again failed to converge, resulting in a WER of 100\% on this dataset, despite having a significantly larger number of parameters and higher computational cost than other approaches. ViT, another global attention-based model, performs second worst, with a CER of 10.57\% and a WER of 37.35\%, although it has the smallest model size and lowest computational cost.

Furthermore, we analyze the errors made by the base version of our model on the WI split of our sentence-based dataset. As shown in Appendix~\ref{app:substitution_error_matrix}, substitution remains the primary error type, although deletions occur more frequently than in the word-based dataset. In addition to confusion between uppercase and lowercase versions of the same letter, errors caused by extreme data imbalance are more prominent, especially for "Q". Some characters are also frequently misrecognized as "Q" due to its low frequency of occurrence, as illustrated in Appendix~\ref{app:distribution_character_length}.

\subsection{Ablation Study}

In this section, we conduct an ablation study on our word-based dataset. Since our method builds upon CLDNN, we evaluate the performance improvements achieved through a series of incremental changes, including an optimized training procedure, data augmentation, architectural enhancements, and model size scaling.

The CLDNN architecture consists of three convolutional layers with 512, 256, and 128 output channels and kernel sizes of 5, 3, and 3 respectively, each followed by batch normalization, ReLU activation, max pooling, and dropout. These convolutional features are then processed through two BiLSTM layers with hidden size of 64, which connect to a final dense layer with ReLU activation. The model is trained using CTC loss and the Adam optimizer with an initial learning rate of 0.01, while a learning rate scheduler monitors validation loss and reduces the learning rate by a factor of 0.8 with patience of 10 epochs. Data preprocessing uses z-score normalization to standardize the input features.

\begin{table}[ht]
    \centering
    \addtolength{\tabcolsep}{3pt}
    \caption{\textbf{Ablation Study Results.} We evaluate the performance of our model on the WI split of our dataset, incorporating incremental modifications based on CLDNN.}
    \begin{tabular}{lcccc}
        \toprule
        Modifications                 & CER   & WER   & \#Params & MACs \\
        \midrule
        CLDNN                         & 19.05 & 60.20 & 0.75M    & 291M \\
        $+$ Improved training         & 17.81 & 56.54 & 0.75M    & 291M \\
        $+$ Data augmentation         & 17.22 & 55.60 & 0.75M    & 291M \\
        $+$ Reverse dimension order   & 16.09 & 53.62 & 0.92M    & 214M \\
        $+$ Embedding layer           & 19.04 & 58.52 & 0.64M    & 99M  \\
        $+$ Separable convolution     & 18.00 & 56.59 & 0.55M    & 81M  \\
        $+$ Multi-scale convolution   & 18.15 & 56.56 & 0.55M    & 81M  \\
        $+$ Instance normalization    & 15.60 & 51.31 & 0.55M    & 81M  \\
        $+$ GELU                      & 15.34 & 50.88 & 0.55M    & 81M  \\
        $-$ Hidden layer (Ours-Small) & 15.29 & 50.70 & 0.53M    & 79M  \\
        $+$ Scale up (Ours-Base)      & 9.44  & 32.17 & 3.89M    & 600M \\
        \bottomrule
    \end{tabular}
    \label{tab:ablation}
\end{table}

As shown in Table~\ref{tab:ablation}, we first employ an improved training pipeline and data augmentation, which together improve performance by 9.6\% and 7.6\% in CER and WER, respectively, without any modifications to the model architecture. Reversing the dimension order of each stage from 512, 256, 128 to 128, 256, 512 slightly increases model size but reduces both error rates and computational cost. Using embedding layers with smaller dimensions of 64, 128, and 256 increases the CER from 16.09\% to 19.04\% and the WER from 53.62\% to 58.52\%, but reduces the number of parameters and MACs by 30.4\% and 53.7\%, respectively.

A combination of micro modifications to the convolutional blocks, including inverted-bottleneck separable depthwise convolution, multi-scale convolution, instance normalization, and GELU, significantly improves performance to 15.34\% CER and 50.88\% WER, representing gains of 19.4\% and 13.1\%, respectively. Among these changes, instance normalization provides additional improvements of 14.0\% in CER and 9.3\% in WER, building on the gains already achieved by depthwise-dilated separable convolution and multi-scale convolution. This highlights the effectiveness of normalization in projecting representations from different handwriting styles into a shared space, enhancing the model ability to adapt to diverse handwriting patterns.

In addition to modifications to the convolutional encoder, we removed the hidden layers from CLDNN, resulting in the small version of our model. It achieves 15.29\% CER and 50.70\% WER, which are 19.7\% and 15.8\% better than CLDNN, while using only 0.53M parameters and 79M MACs, representing reductions of 29.3\% and 72.9\% compared to CLDNN. We then scale up the model by using three convolutional blocks per stage, encoder dimensions of 128, 256, and 512, a BiLSTM with 128 hidden size, and three BiLSTM layers. This results in the base version of our model, which achieves 9.44\% CER and 32.17\% WER, representing 50.4\% and 46.6\% improvements over CLDNN, using 3.89M parameters and 600M MACs, which are 5.2$\times$ and 2.1$\times$ those of CLDNN.

\section{Discussion}
\subsection{Efficiency}

Across all evaluations, we demonstrate the strong efficiency of our models, particularly the small version, which is designed for deployment on edge devices. It achieves significantly lower error rates compared to other IMU-based HWR approaches while using only 53.3\% of the parameters and 27.1\% of the computational cost of CLDNN. This efficiency can greatly enhance the user experience on capable edge devices by providing better recognition without incurring communication latency from remote servers.

The base version, in contrast, is designed for server-side deployment, prioritizing higher recognition accuracy by processing raw signals with more parameters. Our base-version model achieves the best performance on the WI splits while maintaining comparable model size and computational cost, demonstrating a strong balance between efficiency and performance.

\subsection{Robustness}

Handwriting styles vary greatly among individuals, posing significant challenges for HWR. Since it is unrealistic to train a model on data that captures every possible handwriting style, evaluating models on WI datasets provides a more practical and meaningful measure of performance. WI evaluation ensures that handwriting styles in the training and test sets do not overlap, more accurately reflecting real-world scenarios where a model must generalize to unseen writers. Our models outperform all competitors across all WI splits, demonstrating a strong ability to extract robust handwriting features that generalize effectively across diverse writing styles.

Children are beginners in handwriting, and their writing styles differ significantly from those of adults, making HWR more challenging. However, children are also an important user group for HWR systems, given the frequent handwriting activities in educational settings. Therefore, developing a solution that performs well for both adults and children is essential for the success of an HWR system. Our robustness evaluation shows that our models outperform all competitors on both adult and children subsets. Moreover, the results from the cross-subset evaluation suggest that our models learn more generalizable representations, as they maintain better performance even when trained and validated on different age groups.

Additionally, noise caused by various properties of the writing surface, such as hardness and roughness, and sensor imperfections is common in HWR systems and can potentially degrade performance, limiting the environments in which these systems can be used. In our ablation study, we demonstrate that using data augmentations such as AddNoise, Drift, and Dropout can significantly improve the performance of our HWR system. This shows that data augmentation can be an effective solution for handling noise and improving the system robustness.

\subsection{Flexibility}

Due to the nature of handwriting, handwriting signals always vary in length. Models like ViT and Swin Transformers, constrained by positional embeddings or shifting mechanisms, require suboptimal compromises such as padding or truncating to process inputs of different sizes. However, padding increases meaningless computational overhead, while truncating can lead to the loss of important information. In contrast, the CNN-BiLSTM design processes inputs of any size without resizing, enabling efficient computation and adaptability. Thus, approaches that can process inputs of varying sizes are essential for IMU-based HWR.

\section{Conclusion \& Outlook}

In conclusion, our experiments demonstrate that the proposed models consistently outperform competitors on WI datasets, showcasing strong robustness, efficiency, and flexibility in HWR.

However, we have not yet investigated the error patterns in the data. By identifying and addressing semantic errors such as typographical errors or misalignments between sensor signals and text labels, we could curate the datasets to improve their quality. This refinement could enhance HWR system performance, produce more reliable results, and offer deeper insights into the remaining challenges in HWR.

Additionally, although the models show strong performance on our sentence-based dataset, character-level recognition remains a suboptimal approach for leveraging contextual information. It results in a pattern-matching solution that lacks understanding of language semantics from the sensor signals. This limitation could potentially be addressed by incorporating natural language processing techniques into HWR systems.

\clearpage
\appendix
\section{Visualization of Character and Length Distribution}
\label{app:distribution_character_length}

\begin{figure}[htbp]
    \centering
    \includegraphics[width=\textwidth]{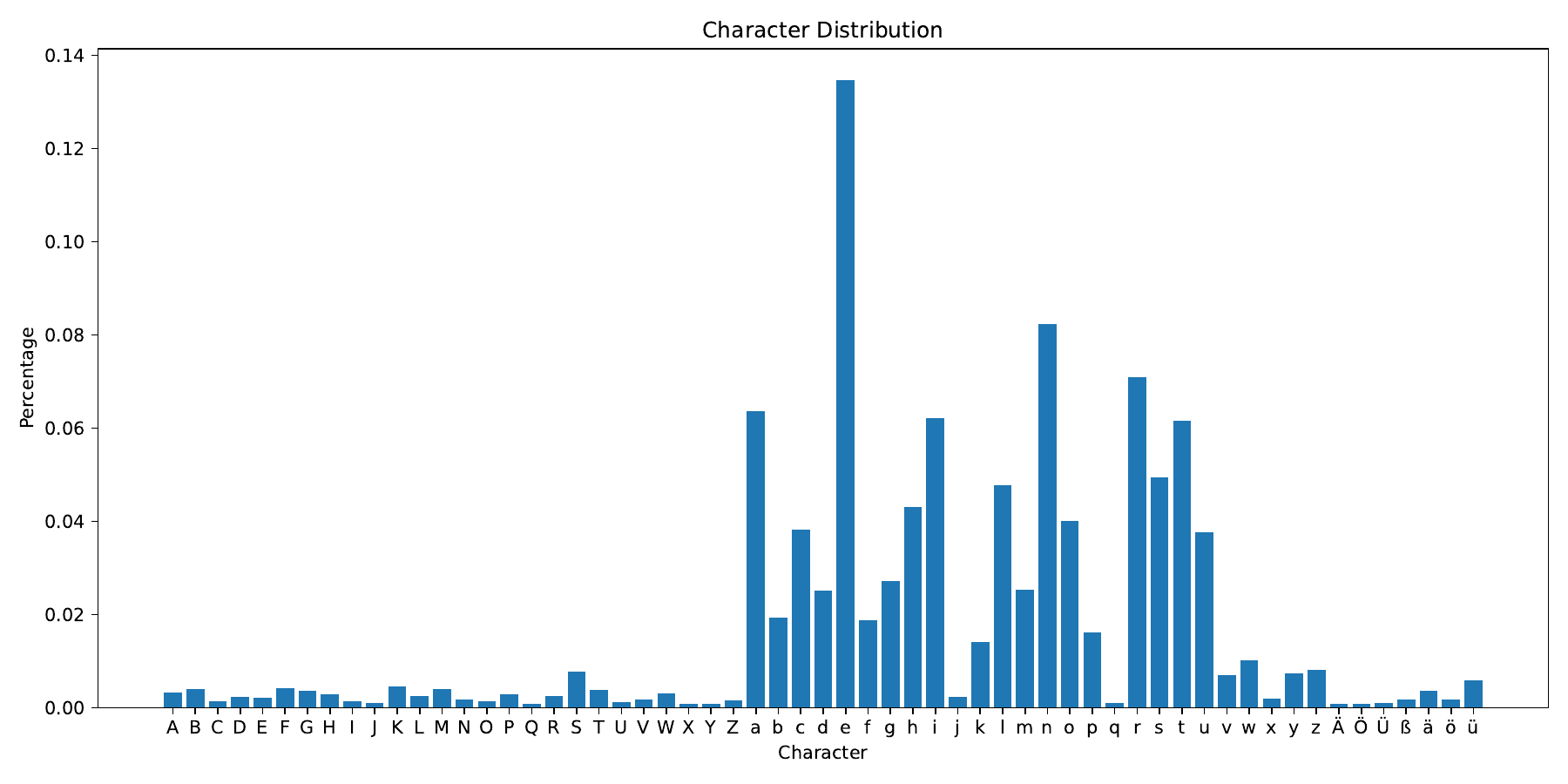}
    \includegraphics[width=0.5\textwidth]{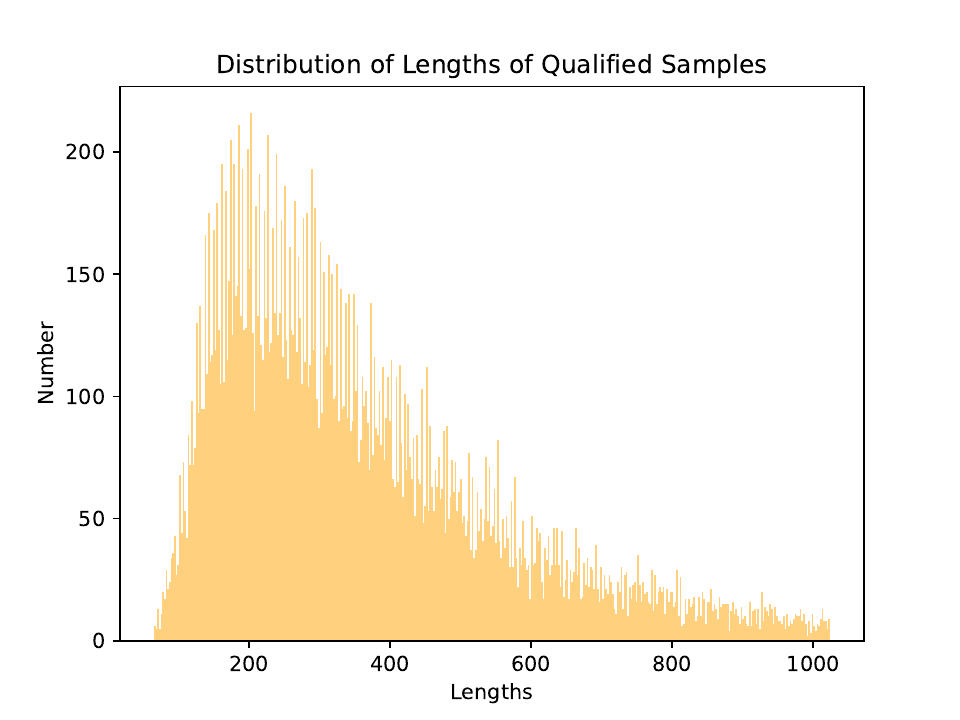}
    \caption{\textbf{Character and sequence length distributions in the word-based dataset.} The top plot shows the frequency of each character, while the bottom plot shows the distribution of sequence lengths.}
    \label{fig:distribution_word}
\end{figure}

\begin{figure}[htbp]
    \centering
    \includegraphics[width=\textwidth]{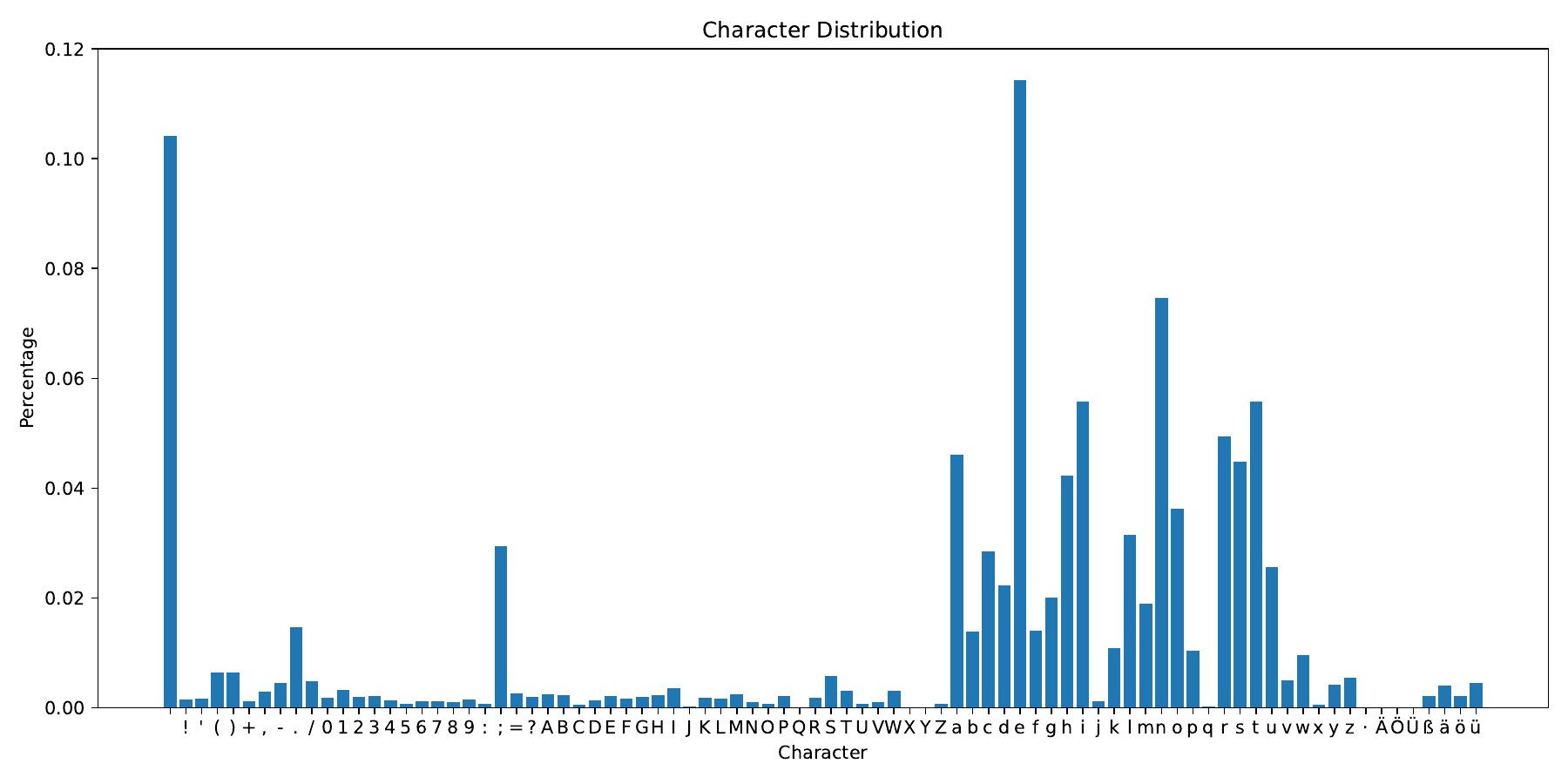}
    \includegraphics[width=0.5\textwidth]{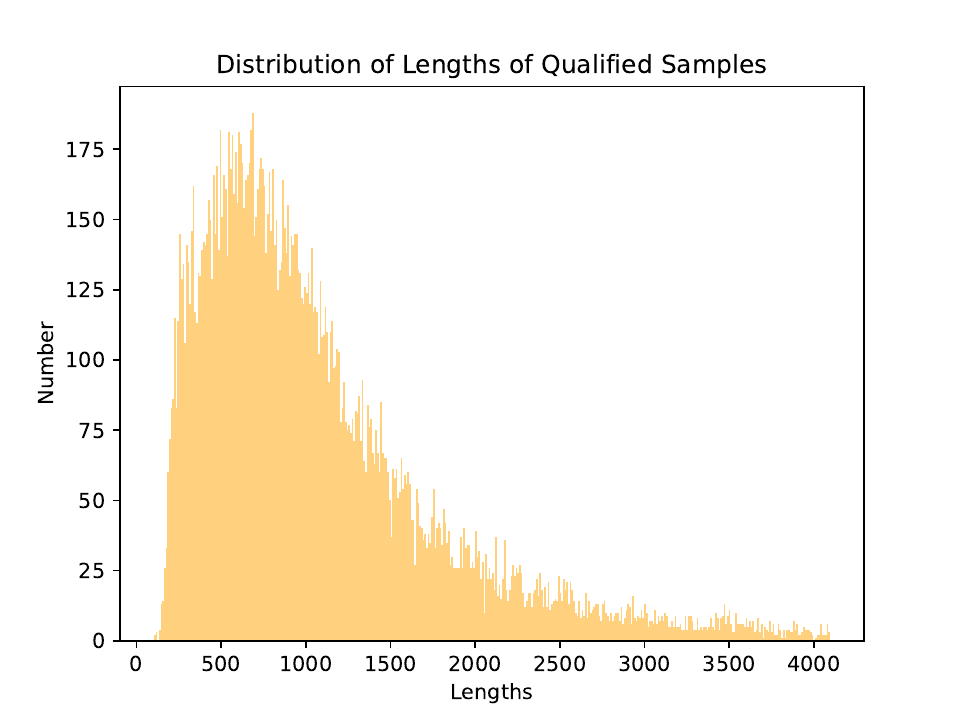}
    \caption{\textbf{Character and sequence length distributions in the sentence-based dataset.} The top plot shows the frequency of each character, while the bottom plot shows the distribution of sequence lengths.}
    \label{fig:distribution_sent}
\end{figure}

The character distributions show consistent patterns across our datasets. Both word-based (Figure~\ref{fig:distribution_word}) and sentence-based (Figure~\ref{fig:distribution_sent}) datasets are dominated by lowercase letters, with "e", "n", and "r" appearing most frequently. This matches the typical letter frequencies in English and German. In the sentence-based dataset, the space character ranks as the second most frequent character, which makes sense given its essential role in sentence structure.

\begin{figure}[htbp]
    \centering
    \includegraphics[width=\textwidth]{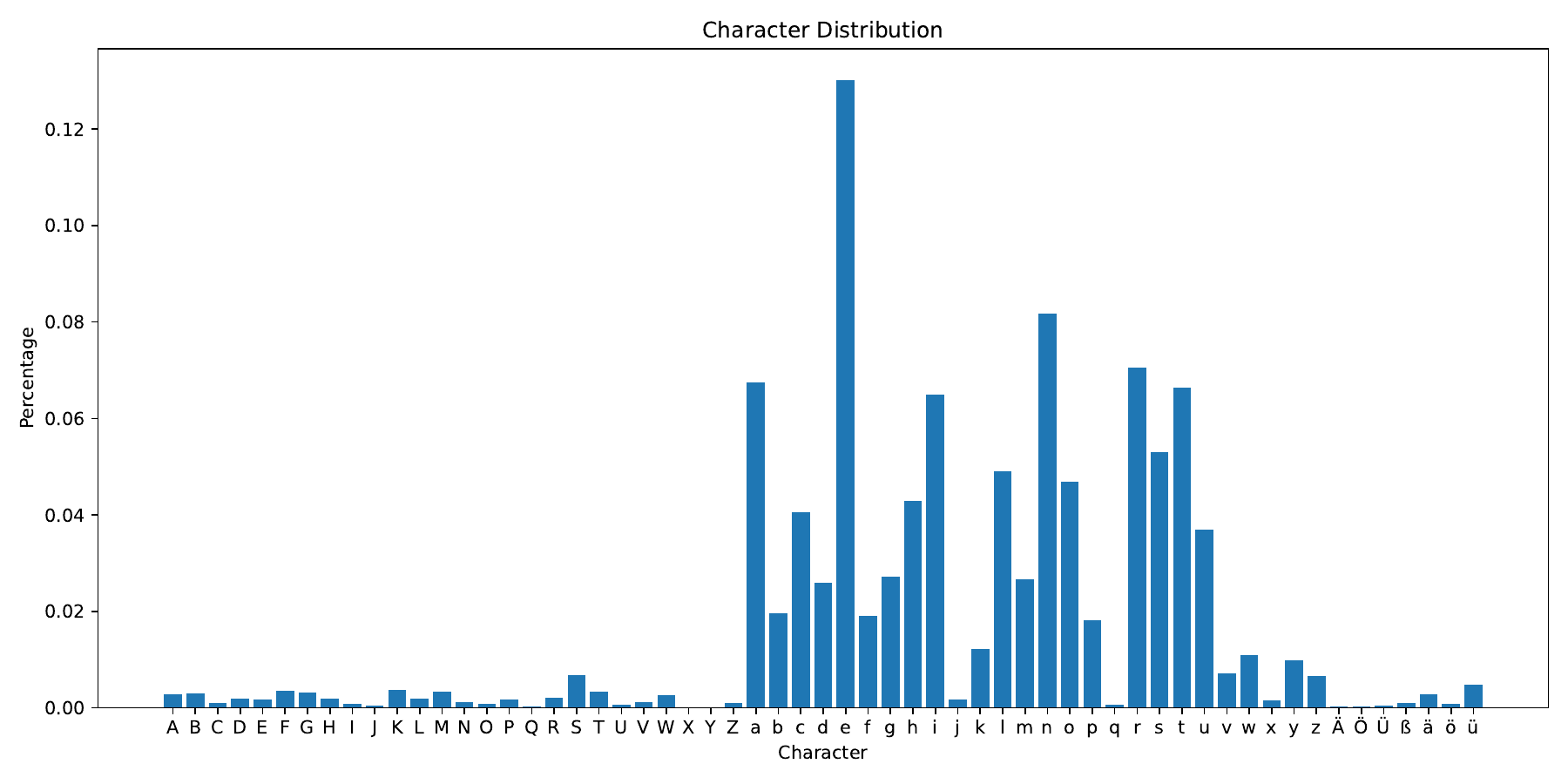}
    \includegraphics[width=0.5\textwidth]{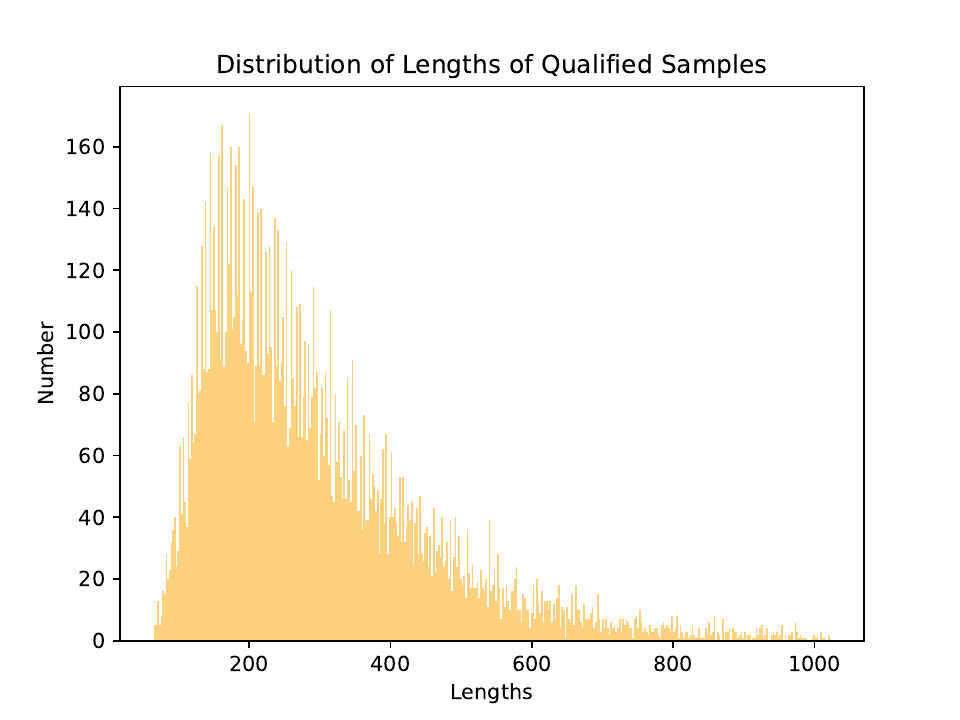}
    \caption{\textbf{Character and sequence length distributions in the adult subset of the word-based dataset.} The top plot shows the frequency of each character, while the bottom plot shows the distribution of sequence lengths.}
    \label{fig:distribution_word_adult}
\end{figure}

\begin{figure}[htbp]
    \centering
    \includegraphics[width=\textwidth]{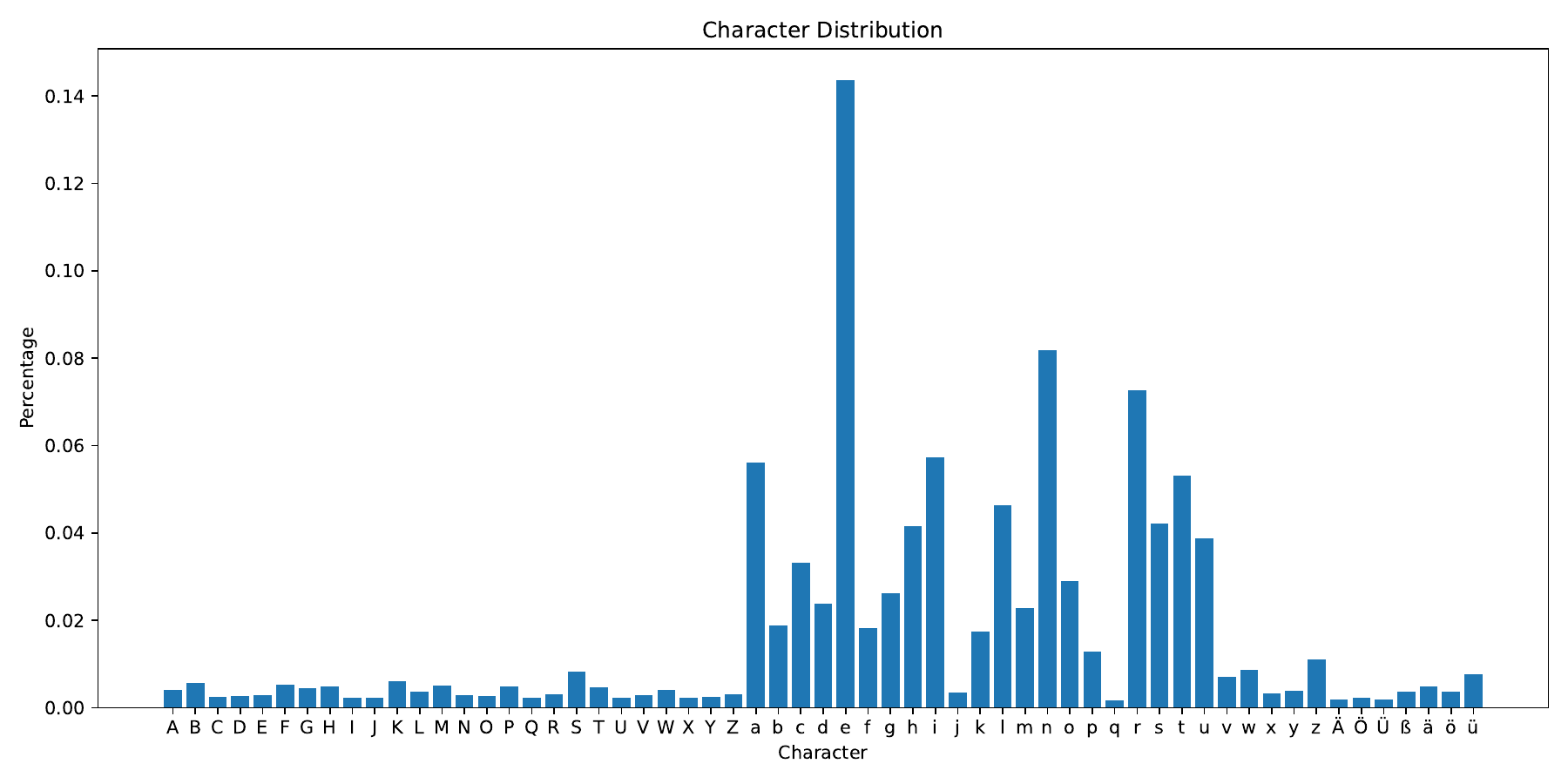}
    \includegraphics[width=0.5\textwidth]{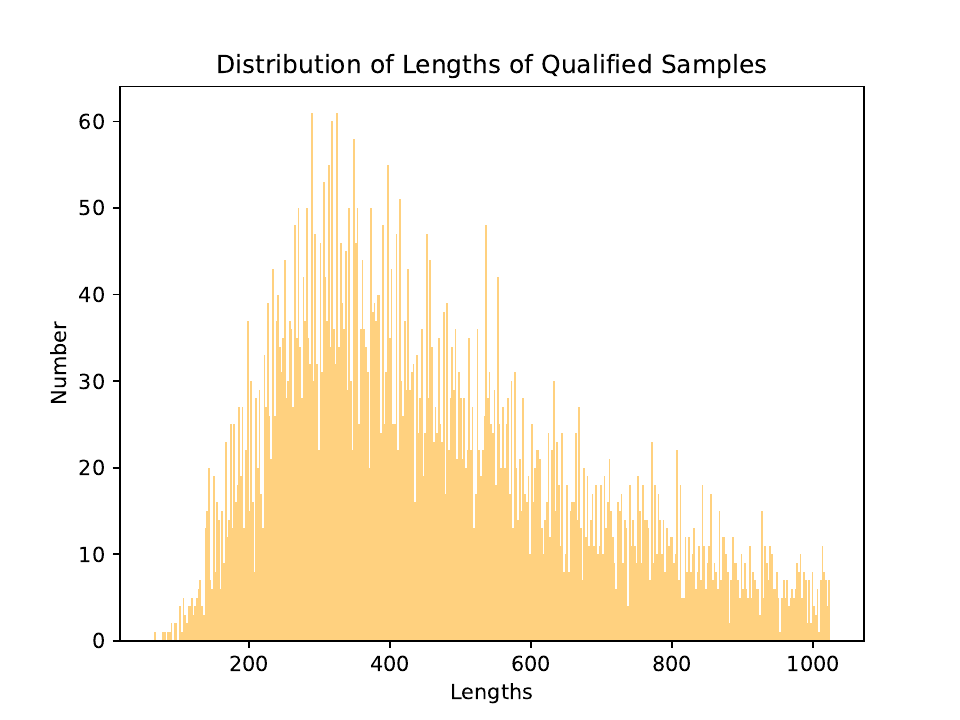}
    \caption{\textbf{Character and sequence length distributions in the children subset of the word-based dataset.} The top plot shows the frequency of each character, while the bottom plot shows the distribution of sequence lengths.}
    \label{fig:distribution_word_child}
\end{figure}

The word-based dataset reveals clear differences between age groups. Adult (Figure~\ref{fig:distribution_word_adult}) and children (Figure~\ref{fig:distribution_word_child}) handwriting samples show distinct distribution patterns, with children samples having longer average lengths than adult samples.

\section{Visualization of Data Augmentation}
\label{app:data_augmentation}

\begin{figure}[htbp]
    \centering
    \includegraphics[width=\linewidth]{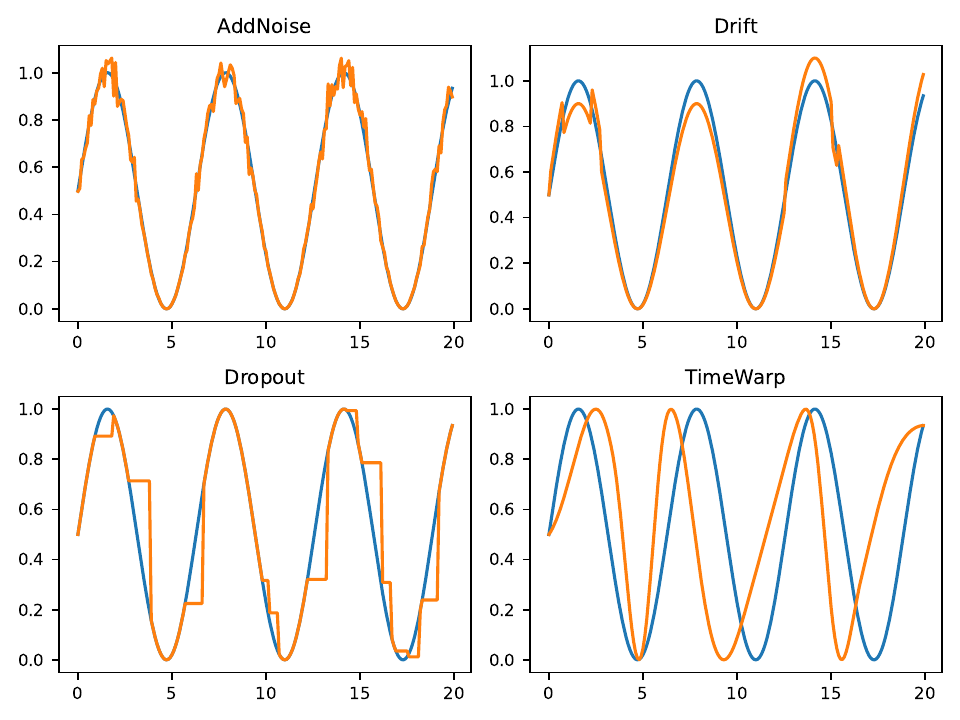}
    \caption{\textbf{Data augmentation.} Examples of sine signals with different data augmentation techniques applied. The blue curve represents the original sine signal and the orange curve represents the augmented signal.}
    \label{fig:data_augmentation}
\end{figure}

As shown in the visualization of data augmentations in Figure~\ref{fig:data_augmentation}, AddNoise, Drift, and Dropout are applied in a multiplicative manner to ensure that the augmentation magnitude neither overshadows the original signal nor dominates it.

\section{Visualization of Substitution Error Rates}
\label{app:substitution_error_matrix}

\begin{figure}[htbp]
    \centering
    \includegraphics[width=\linewidth]{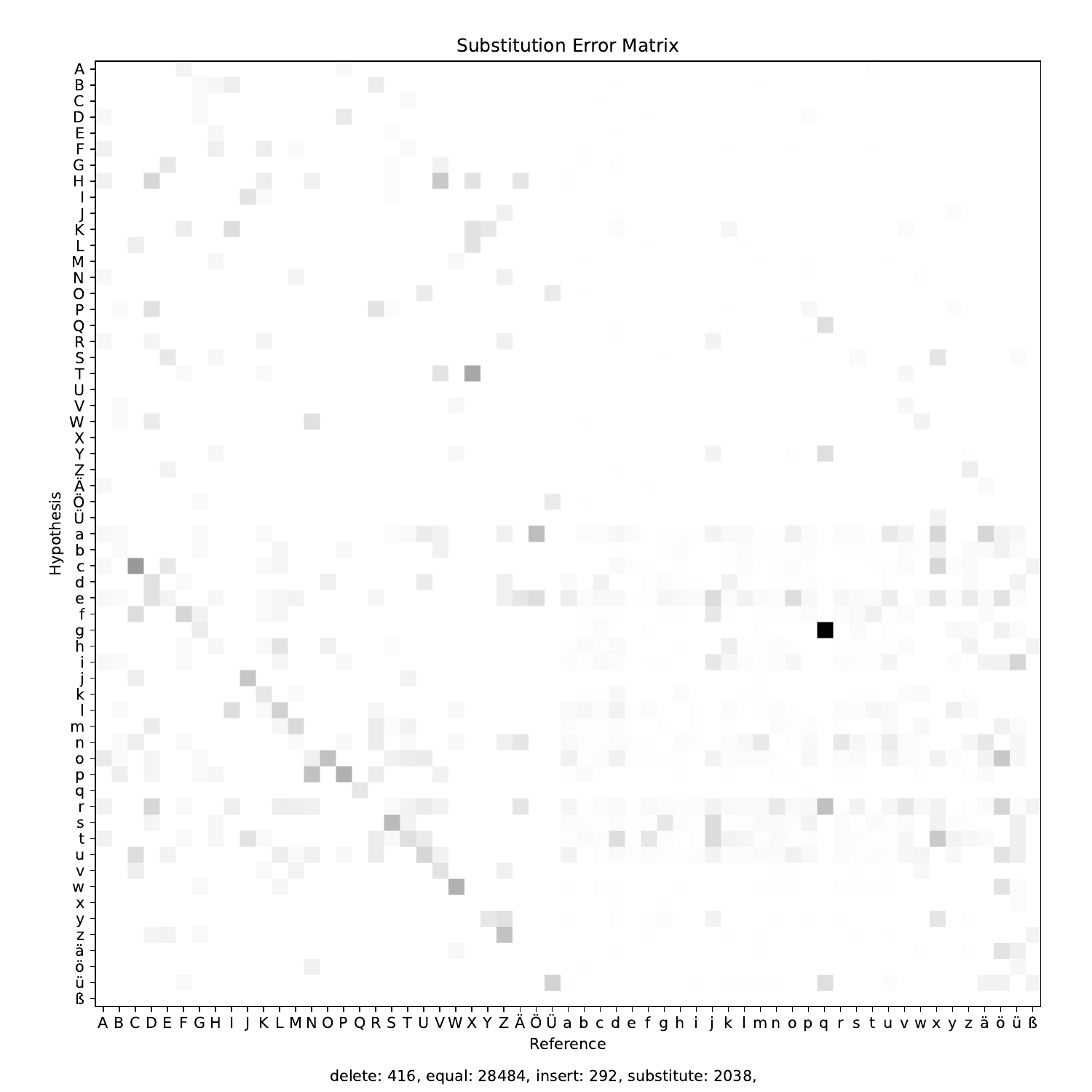}
    \caption{\textbf{Substitution Errors on Word-Based Dataset.} Substitution errors made by the base version of our model on the first fold of the WI split of our word-based dataset are visualized. Darker shades indicate higher error rates.}
    \label{fig:substitution_error_word}
\end{figure}

\begin{figure}[htbp]
    \centering
    \includegraphics[width=\linewidth]{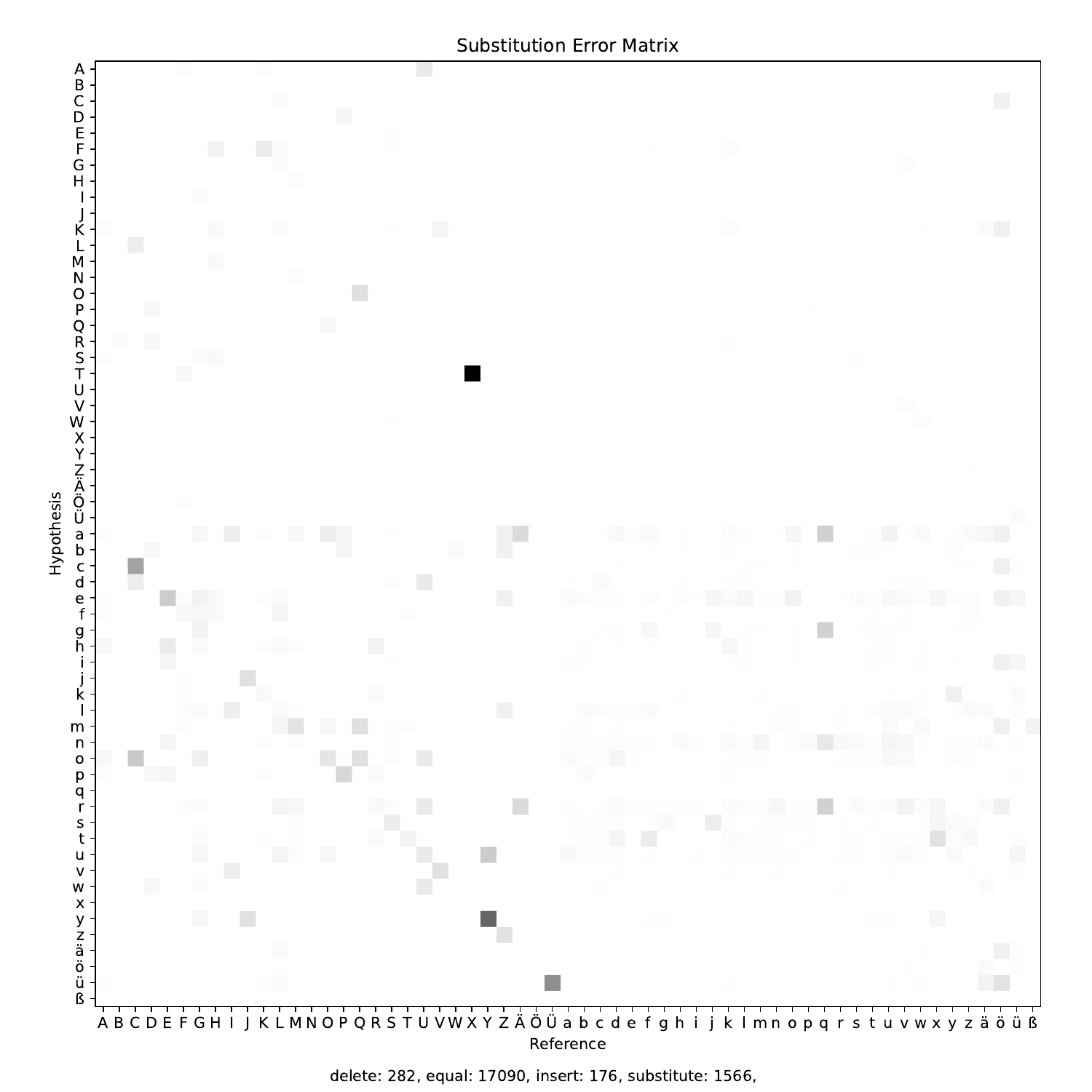}
    \caption{\textbf{Substitution Errors on the Adult Subset.} Substitution errors made by the base version of our model on the first fold of the WI split of the adult subset of our word-based dataset are visualized. Darker shades indicate higher error rates.}
    \label{fig:substitution_error_adult}
\end{figure}

\begin{figure}[htbp]
    \centering
    \includegraphics[width=\linewidth]{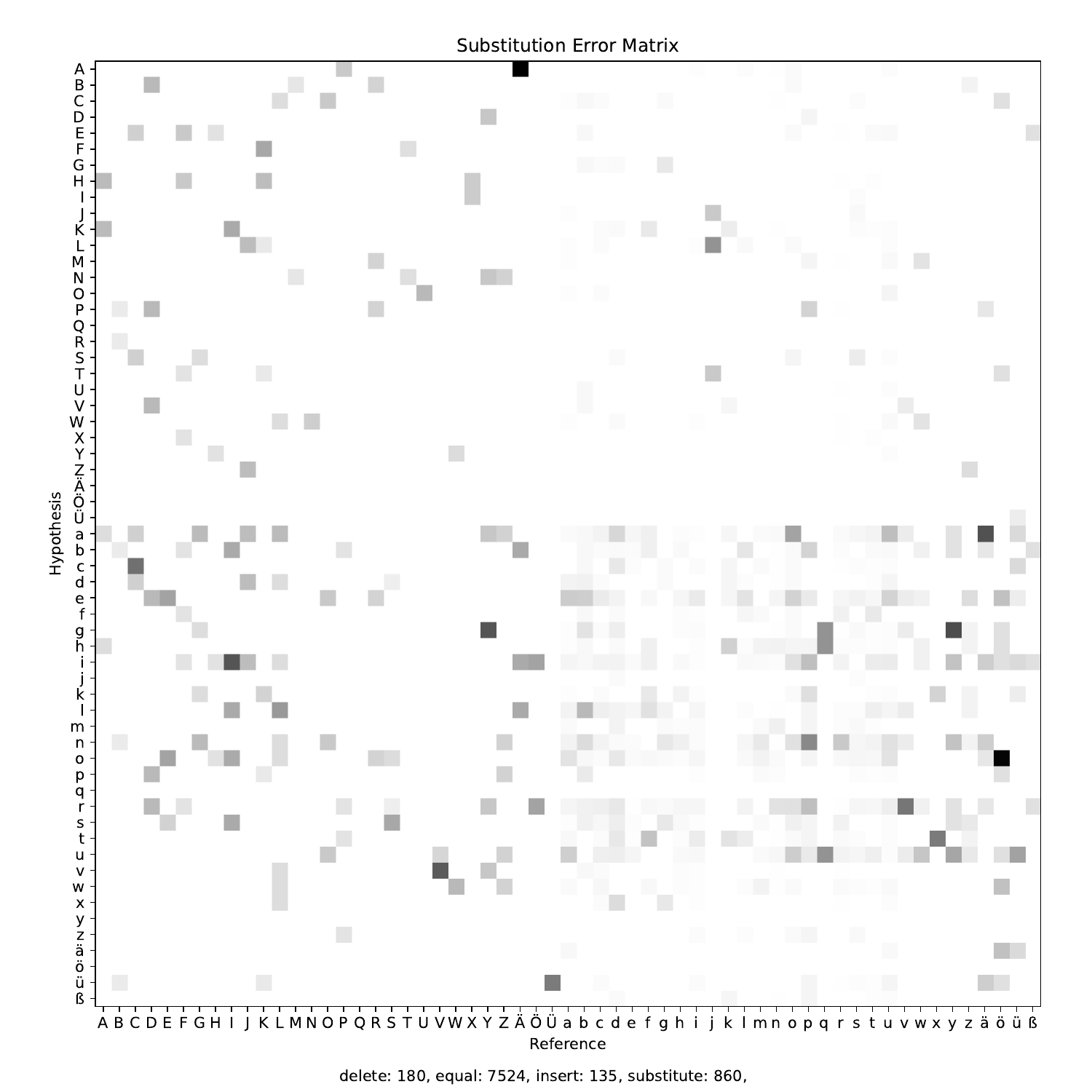}
    \caption{\textbf{Substitution Errors on the Children Subset.} Substitution errors made by the base version of our model on the first fold of the WI split of the children subset of our word-based dataset are visualized. Darker shades indicate higher error rates.}
    \label{fig:substitution_error_child}
\end{figure}

\begin{figure}[htbp]
    \centering
    \includegraphics[width=\linewidth]{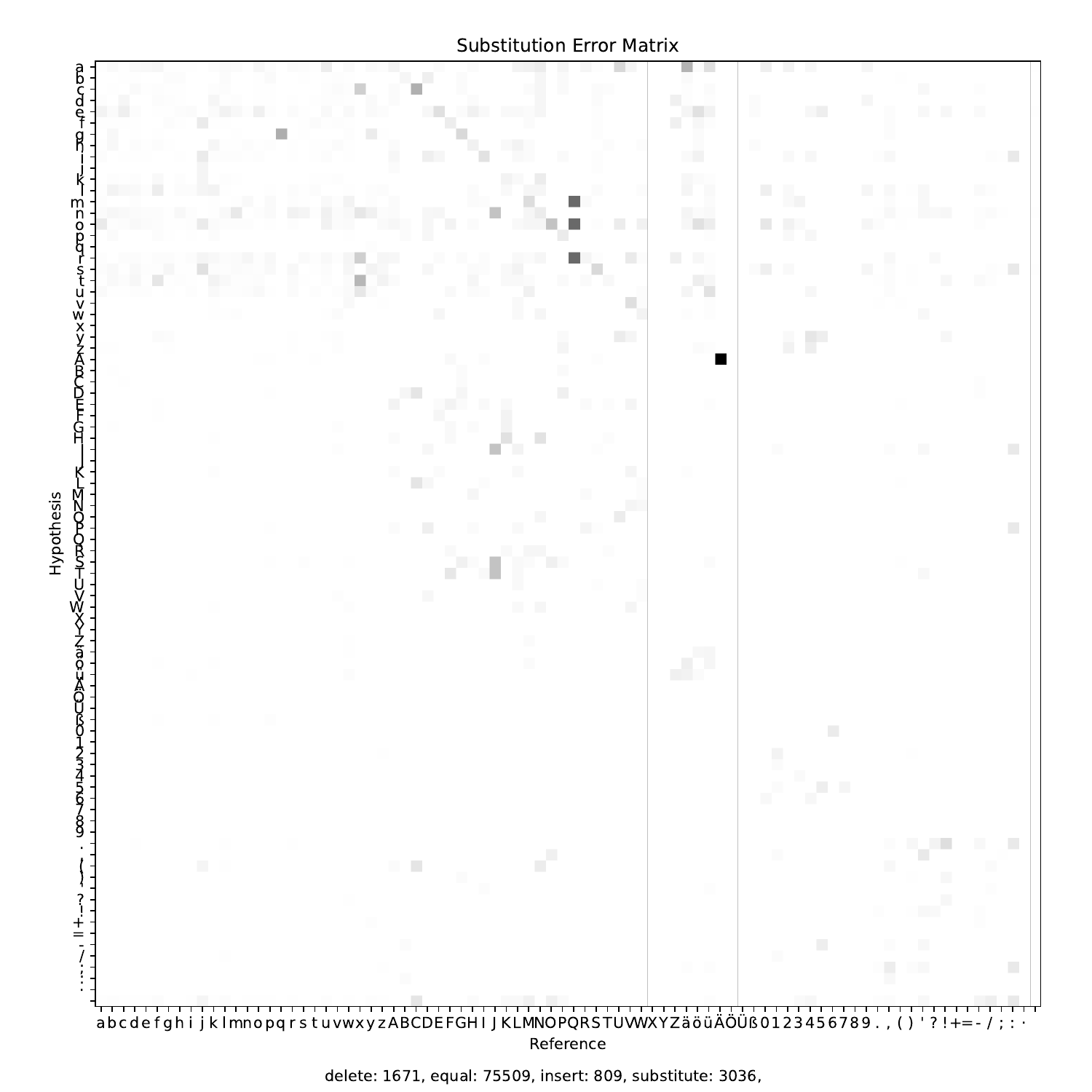}
    \caption{\textbf{Substitution Errors on Sentence-Based Dataset.} Substitution errors made by the base version of our model on the first fold of the WI split of our sentence-based dataset are visualized.}
    \label{fig:substitution_error_sent}
\end{figure}

Figure~\ref{fig:substitution_error_word} shows the substitution error matrix of our base-version model on the first fold of the WI split of our word-based dataset. Figure~\ref{fig:substitution_error_adult} shows the substitution error matrix of our base-version model on the first fold of the WI split of the adult subset of our word-based dataset. Figure~\ref{fig:substitution_error_child} shows the substitution error matrix of our base-version model on the first fold of the WI split of the children subset of our word-based dataset. Figure~\ref{fig:substitution_error_sent} presents the substitution error matrix of the same model on the first fold of the WI split of our sentence-based dataset.

\newpage
%
%
\bibliographystyle{splncs04}
\bibliography{references}

\end{document}